\PassOptionsToPackage{table}{xcolor}
\documentclass{article}
\usepackage[T1]{fontenc}
\usepackage{iclr2027_conference,times}
\usepackage{xcolor}

\usepackage{amsmath}

\usepackage{hyperref}
\hypersetup{
  hidelinks,
  pdftitle={KINESYS: Test-Time Adaptation for Reference-Based Effect Editing in Images},
  pdfauthor={Noah Frahm, Luchao Qi, Haoze Li, Dan B. Goldman, Sambit Bhattacharya, Roni Sengupta}
}
\usepackage{url}
\usepackage{graphicx}
\usepackage{float}
\usepackage{booktabs}
\usepackage{tabularx}
\usepackage{listings}

\lstdefinestyle{prompt}{
  basicstyle=\ttfamily\footnotesize,
  breaklines=true,
  breakatwhitespace=true,
  columns=fullflexible,
  keepspaces=true,
  frame=single,
  framerule=0.25pt,
  framesep=4pt,
  framexleftmargin=0pt,
  framexrightmargin=0pt,
  linewidth=\columnwidth,
  xleftmargin=0pt,
  xrightmargin=0pt,
  resetmargins=true,
  aboveskip=6pt,
  belowskip=8pt,
  showstringspaces=false
}

\title{KINESYS: Test-Time Adaptation for\\
Reference-Based Effect Editing in Images}

\author{%
  \textbf{Noah Frahm\textsuperscript{1},
  Luchao Qi\textsuperscript{1}\thanks{Denotes equal contribution.},
  Haoze Li\textsuperscript{1}\footnotemark[1]} \\
  \textbf{Dan B Goldman\textsuperscript{2},
  Sambit Bhattacharya\textsuperscript{3},
  Roni Sengupta\textsuperscript{1}} \\[0.5ex]
  \textsuperscript{1}University of North Carolina at Chapel Hill \\
  \textsuperscript{2}Industrial Light \& Magic \\
  \textsuperscript{3}Fayetteville State University
}

\iclrfinalcopy
\begin{document}
\pagenumbering{arabic}
\maketitle

% Neutral arXiv page style; the title macro sets a conference header.
\fancyhf{}
\fancyfoot[C]{\thepage}
\renewcommand{\headrulewidth}{0pt}
\renewcommand{\footrulewidth}{0pt}
\pagestyle{fancy}
\thispagestyle{fancy}

\begin{figure}[h]
\centering
\includegraphics[width=0.94\textwidth,trim=42mm 42mm 42mm 42mm,clip]{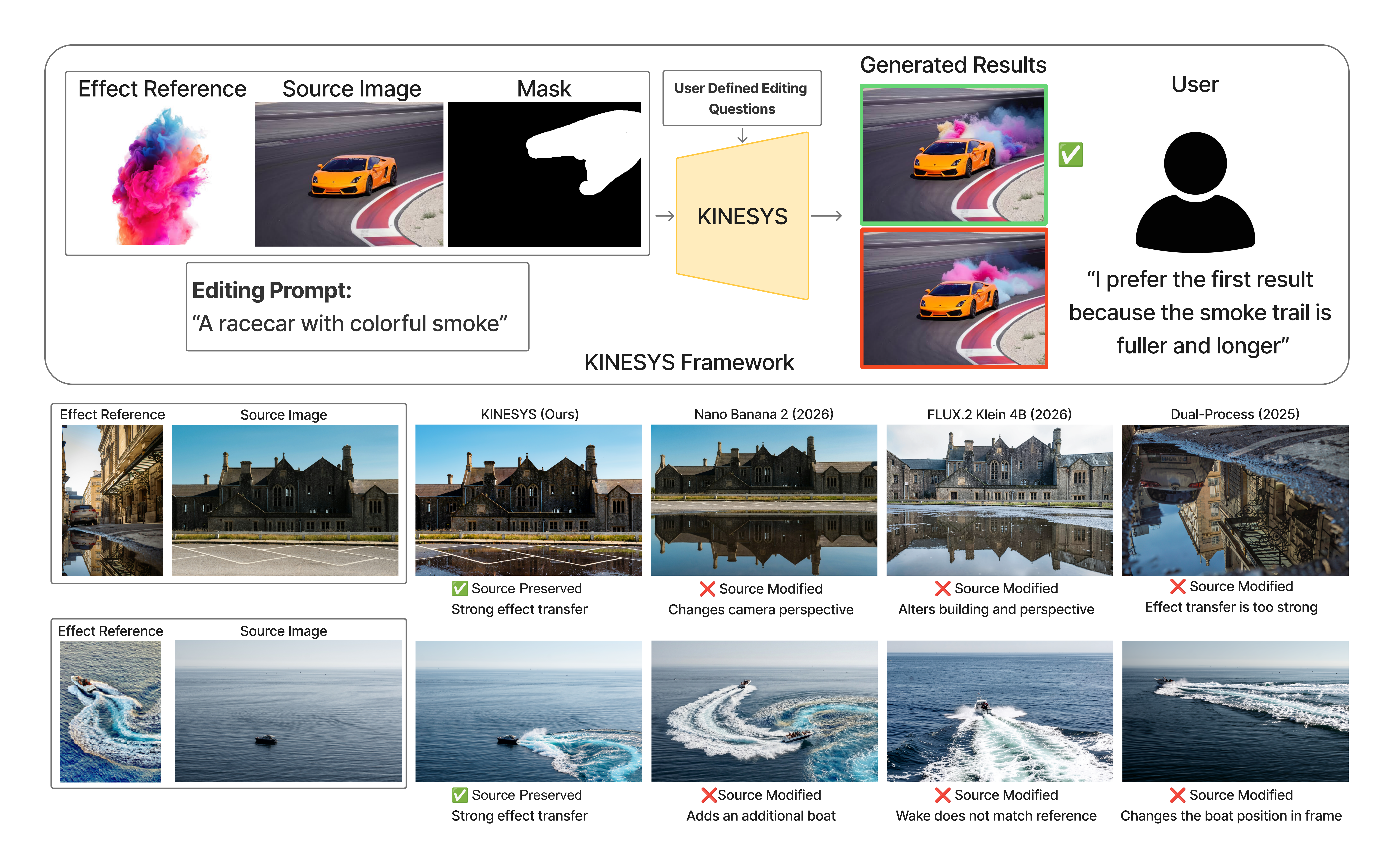}
\par\vspace{-0.5em}
\caption{We present \textsc{Kinesys}, a test-time adaptation framework for reference-based visual effects editing in images. (Top) The artist supplies a source, effect reference, optional mask, editing prompt, and verification questions, and \textsc{Kinesys} generates multiple samples with diverse styles. (Bottom) \textsc{Kinesys} transfers puddle and boat-wake effects while preserving source composition and content; baselines change the scene, add unwanted content, or miss effect properties.}

\label{fig:main_teaser}
\vspace{-0.5em}
\end{figure}

\begin{abstract}
Visual effects are often easier to demonstrate via reference than to describe in words. We study effect-reference editing, where a reference image shows a desired visual effect and language identifies which properties should guide the edit. Existing single-pass editing techniques often fail to make fine-grained edits, resulting in unintended changes to source content or deviations from the reference effect. We introduce \textsc{Kinesys}, a test-time adaptation framework for effect-reference editing that combines reference-aware and source-aware feedback to transfer the desired effect while preserving the composition and content of the original shot. To ensure \textsc{Kinesys} can produce a diverse set of edits that preserve the source, we develop a robust optimization strategy in which the generative model is fine-tuned across multiple noise samples and optimization runs. Evaluation with VLM judges, source preservation metrics (LPIPS, TPIPS), and a user study shows that \textsc{Kinesys} outperforms state-of-the-art image editors in source preservation; on effect transfer, it remains competitive with Nano Banana 2 and outperforms open-weight editors.
\quad\textbf{Project website:} \href{https://noahfrahm.github.io/KINESYS-Project-Page/}{KINESYS-Project-Page}

\end{abstract}
\section{Introduction}
\label{sec:introduction}
\vspace{-1em}

Recent open-weight and proprietary image editors can make convincing changes to existing images~\citep{flux-2-2025,wu2025qwenimagetechnicalreport,google2026nanobanana2,openai2026gptimage2,seedream2025seedream4}. Editing visual effects nevertheless exposes a difficult control problem: artists often need to add elements such as smoke, fire, or rain, or introduce visible damage, without unintended changes to the source subjects, camera pose, framing, or scene object layout. An instruction-based edit such as ``add tire smoke behind the car'' names the new element but does not specify its texture, color, scale, density, placement, or interaction with the road and vehicle. A visual reference can supply these details, but the editor must separate the properties that define the desired effect from incidental reference content. The central challenge is therefore to realize the requested effect strongly enough while preserving the source subject and scene structure.

We study \emph{effect-reference editing}: the user provides a source image, an effect reference, and a text instruction identifying which visual and spatial properties should guide the edit. The user specifies which effect properties to transfer through reference-grounded yes-or-no questions and may include a mask to localize the edit. The effect reference specifies what to introduce. We use \emph{source preservation} to mean retaining the camera pose, framing, scene object layout, and subject appearance except where the requested effect requires a change. Modern in-context editors can accept source images, references, and language through a common interface~\citep{flux-2-2025,seedream2025seedream4}, making them a natural basis for this setting.

A single editing pass may under-edit, producing a weak or generic result, or follow the reference too closely and alter the subject, viewpoint, or scene layout, as illustrated in Figure~\ref{fig:main_teaser}. A natural response to single-pass failures is to evaluate candidate edits and use that feedback to improve subsequent outputs. \textsc{Kinesys} fine-tunes the generator using reference-grounded visual question answering (VQA): a vision-language model (VLM) answers the supplied questions about each candidate. These judges provide different information: our Gemma-based VQA judge~\citep{gemmateam2026gemma4} sees the reference but not the source and assesses the selected effect properties, while EditReward~\citep{wu2026editrewardhumanalignedrewardmodel} sees the source but not the reference and assesses instruction following and visual quality. We combine these judges to support effect transfer that preserves the source.

However, optimizing the generator with complementary feedback remains challenging. The same editing request can produce different results depending on the random noise used to generate its candidates~\citep{ma2025inferencetimescalingdiffusion,guimaraes2026progressiveseedpruning,benhamu2024dflow,kim2025flowinferencescaling}. Because these candidates also provide the feedback for fine-tuning, an unlucky initial set can steer the generator toward a poor solution. To reduce this dependence, \textsc{Kinesys} starts $K$ independent fine-tuning runs, each using $N$ fixed-noise candidates, then prunes runs and selects a checkpoint.

Together, these design choices adapt \textsc{Kinesys} to one request, allowing it to produce several versions of the referenced effect from unseen noise samples while preserving the source. As Figure~\ref{fig:fresh_noise_variations} shows, new noise changes details such as texture, shape, density, and placement, giving artists several interpretations of the same effect to compare.

We evaluate \textsc{Kinesys} on 81 editing requests spanning eight visual effects.
\textsc{Kinesys} achieves the strongest source preservation scores among the evaluated methods, remains competitive with Nano Banana 2 on editing quality, and leads the evaluated open-weight editors on both source preservation and editing quality. The blinded user study shows that participants selected \textsc{Kinesys} more often for source preservation against Nano Banana 2~\citep{google2026nanobanana2} ($45.3\%$ versus $24.7\%$) and Step1X-Edit~\citep{liu2025step1xedit} ($53.6\%$ versus $20.9\%$), while effect-transfer preferences were competitive with Nano Banana 2 and favored \textsc{Kinesys} over Step1X-Edit.

Our contributions are threefold:
\begin{itemize}
    \item We introduce \textsc{Kinesys}, a test-time generator adaptation framework for reference-based visual effect editing in images
    that uses reference-grounded and source-aware VLM feedback to perform effects editing that preserves the source without effect-specific training data.
    \item We develop a robust optimization strategy for fine-tuning the generator with LoRA using $N$ fixed-noise candidates per run and pruning across $K$ independent runs, with an early diversity penalty to discourage candidate collapse.
    \item We create a benchmark of 81 editing requests across eight effects with VLM and human judges measuring editing quality for effect transfer and source preservation. Our method outperforms the evaluated open-weight models and surpasses Nano Banana 2 in source preservation while remaining competitive on effect transfer.
\end{itemize}

\section{Related Work}
\label{sec:related_work}

\noindent\textbf{Dataset-trained and reference-guided image editing.}\enspace
Image editors learn shared behavior from paired or synthetic data, as in InstructPix2Pix~\citep{brooks2023instructpix2pixlearningfollowimage}. Paint by Example and MimicBrush transfer reference content into source regions~\citep{yang2022paintexampleexemplarbasedimage,chen2024mimicbrush}; FreeEdit combines reference and language for mask-free editing~\citep{he2024freeedit}; and Visual Instruction Inversion infers an operation from a before-and-after pair~\citep{nguyen2023visualinstructioninversionimage}. Recent image editors and unified multimodal models condition image generation or editing on images and language~\citep{labs2025flux1kontextflowmatching,flux-2-2025,wu2025qwenimagetechnicalreport,zhang2025icedit,seedream2025seedream4,jiang2026orionedit,diao2026sensenovau1}. \textsc{Kinesys} adapts the generator to one effect-reference request without an effect-specific training set or target edited image.

\noindent\textbf{Generative visual effects.}\enspace
Recent visual-effects generation and compositing work includes MagicVFX, VFX Creator, Omni-Effects, and Over++~\citep{guo2024magicvfx,liu2025vfxcreator,mao2026omnieffects,qi2025overplusplus}. EffectMaker learns reference-based VFX generation from a video dataset, while P-Flow refines prompts against a reference at test time with a fixed video generator~\citep{yang2026effectmaker,zhao2026pflow}. \textsc{Kinesys} instead adapts a generator for one image edit using a source and effect reference, with the reference defining the effect and the source \mbox{defining what stays unchanged.}

\noindent\textbf{Request-specific and feedback-driven adaptation.}\enspace
Imagic and UniTune~\citep{kawar2023imagictextbasedrealimage,51891} fine-tune a diffusion model for one source image and target text. DRaFT directly optimizes diffusion parameters against a reward, while HIVE, InstructRL4Pix, CogniEdit, and NP-Edit train shared editors with feedback or dataset supervision~\citep{clark2024directlyfinetuningdiffusionmodels,zhang2023hive,li2024instructrl4pixtrainingdiffusionimage,li2026cogniedit,kumari2026npedit}. SPIE uses AI feedback for concept-specific adaptation from a small set of reference images~\citep{benarous2025spie}.

Most closely related, Dual-Process uses a VLM to guide request-specific fine-tuning of an image generator~\citep{luo2025dualprocessimagegeneration}. It asks whether generated images satisfy a control question, then uses the answer likelihood to update LoRA weights on the generator. This provides a way to learn a visual control without a task-specific training set. \textsc{Kinesys} extends this VLM-guided adaptation to effect-reference editing, with the source image constraining subject appearance and scene composition.

\noindent\textbf{Complementary feedback for reference-guided editing.}\enspace
HIVE, ADIEE, EditScore, and EditReward evaluate an edit against its source and instruction without an effect reference~\citep{zhang2023hive,chen2025adiee,luo2026editscoreunlockingonlinerl,wu2026editrewardhumanalignedrewardmodel}. Reference-aware evaluators include ICE-Bench, which judges the instruction, source, reference, and result, and EditCLIP, which compares an edit with a paired reference transformation~\citep{pan2025icebench,wang2025editclip}. EVR uses multidimensional reference-aware feedback for dataset-level post-training of multi-reference editors~\citep{miao2026evaluationverificationreward}. \textsc{Kinesys} combines both feedback roles within request-specific fine-tuning.

\noindent\textbf{Reward-guided inference and test-time scaling.}\enspace
Inference-time scaling commonly holds the generator fixed. Best-of-$N$ and adaptive pruning search over noise inputs or partial candidates~\citep{ma2025inferencetimescalingdiffusion,guimaraes2026progressiveseedpruning,qu2026adecot,yu2026verilatent}; D-Flow, RewardFlow, FMTT, trajectory control, and closed-loop generation optimize noise, sampling dynamics, or latent controls~\citep{benhamu2024dflow,susladkar2026rewardflow,sabour2025fmtt,chang2026trajectorycontrol,zhao2026closedloop}; and iterative editors revise outputs over multiple rounds~\citep{bai2026uniediti,zeng2025mira}. \textsc{Kinesys} instead allocates compute across independent fine-tuning runs and returns an editor that can produce several versions of the referenced effect while preserving the source.

\begin{figure}[!t]
    \centering
    \includegraphics[width=\textwidth,trim=42mm 42mm 42mm 42mm,clip]{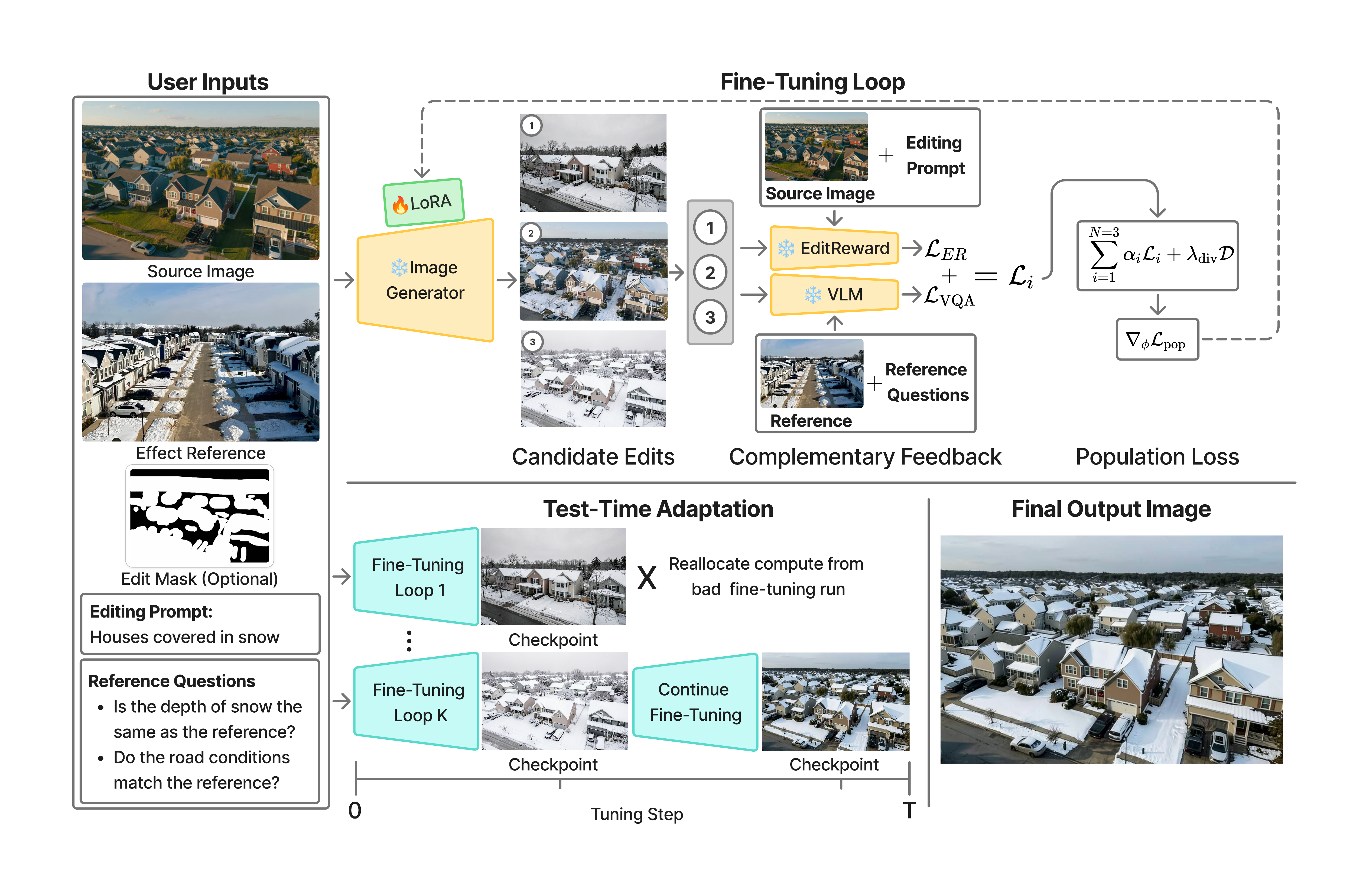}
    \par\vspace{-0.6em}
    \caption{Overview of \textsc{Kinesys}. Reference-grounded VQA and source-aware EditReward provide complementary feedback. Each of $K$ independent runs has its own LoRA parameters $\phi_k$ and $N$ fixed-noise candidates. Within a run, $\alpha_i$ weights candidate $i$'s objective, while the decaying term $\lambda_{\mathrm{div}}(t)\mathcal{D}$ discourages mode collapse. The axis from $0$ to $T$ marks optimization steps; less promising runs stop at pruning points, while survivors continue to later checkpoints.}
    \label{fig:kinesys_architecture}
    \vspace{-1em}
\end{figure}

\section{Method}
\label{sec:method}

An effect reference supplies appearance cues, but it also contains subjects and scene details that should not enter the source. \textsc{Kinesys} addresses this by adapting an image editor for each request using both reference-aware and source-aware feedback. The inputs are a source image, an effect reference, a text instruction, reference-grounded questions, and an optional spatial mask. The reference shows the desired effect on different content; the user chooses which visual and spatial properties to transfer and specifies them through the instruction and reference-grounded questions. Without a mask, the entire source image may be edited. The adapted editor can then produce multiple versions of the effect from different noise samples while preserving the source content.

Existing editors can produce convincing image edits, but on effect-reference requests they can miss selected effect properties or alter the source subject and scene, as Figure~\ref{fig:main_teaser} illustrates. We propose \textsc{Kinesys}, a test-time adaptation strategy where VLM judges evaluate effect transfer and preservation of the source image structure and provide feedback to update the generative model.

At each fine-tuning step, \textsc{Kinesys} renders $N$ candidate edits from different fixed noise samples. Reference-grounded VQA and EditReward evaluate each candidate. \textsc{Kinesys} uses their feedback to update only the LoRA weights $\phi$~\citep{hu2021loralowrankadaptationlarge}. The base editor, text encoder, and feedback models remain fixed. Independent fine-tuning runs explore different noise sets before the final adapted LoRA weights are selected. Figure~\ref{fig:kinesys_architecture} illustrates the \textsc{Kinesys} framework.

\subsection{Complementary Editing Feedback}
\label{sec:method_editor}
\label{sec:method_objective}

Following Dual-Process~\citep{luo2025dualprocessimagegeneration}, a frozen vision-language model compares the effect reference $r$ with each candidate edit $y_i$ and answers targeted questions about properties such as the appearance, placement, or interaction of the introduced element. For each of the $Q$ questions $q_\ell$, the desired answer $a_\ell$ is \texttt{Yes}. The VQA loss averages the negative log-likelihood of these answers, so lower loss indicates clearer realization of the selected effect properties.

VQA compares each candidate edit with the effect reference but does not see the source, so it cannot directly assess unwanted changes to the original subject or scene. We therefore also use EditReward~\citep{wu2026editrewardhumanalignedrewardmodel}, a learned reward model that sees the source, instruction, and candidate edit. It predicts two scores: instruction following $R_{i,\mathrm{follow}}$ and visual quality $R_{i,\mathrm{quality}}$. The first judges whether an edit is accurate, complete, and limited to the request; the second judges realism and visible artifacts. We average these model outputs as $R_{\mathrm{ER},i}:=(R_{i,\mathrm{follow}}+R_{i,\mathrm{quality}})/2$ and minimize $\mathcal{L}_{\mathrm{ER},i}:=-R_{\mathrm{ER},i}$. The per-candidate objective combines VQA and EditReward:

\noindent
\begin{minipage}{0.49\linewidth}
\begin{equation}
    \mathcal{L}_{\mathrm{VQA},i}
    = -\frac{1}{Q}\sum_{\ell=1}^{Q}
      \log p_{\mathrm{VLM}}\!\left(a_\ell \mid r,y_i,q_\ell\right)
    \label{eq:vqa_loss}
\end{equation}
\end{minipage}\hfill
\begin{minipage}{0.49\linewidth}
\begin{equation}
    \mathcal{L}_i
    = \lambda_{\mathrm{VQA}}\mathcal{L}_{\mathrm{VQA},i}
    + \lambda_{\mathrm{ER}}\mathcal{L}_{\mathrm{ER},i}
    \label{eq:task_objective}
\end{equation}
\end{minipage}

Because optimization can exploit blind spots in learned rewards, we use these two complementary objectives: VQA should strengthen transfer of the selected effect, while EditReward should limit unintended source changes and visual artifacts. The ablations in Sec.~\ref{sec:experiments_ablations} and Fig.~\ref{fig:feedback_ablations} \mbox{examine these expectations.}

\subsection{Robust Optimization}
\label{sec:method_optimization}

\textbf{Learning from several candidate edits.}
A single noise sample can produce a weak edit or one with unintended source changes, and fine-tuning only that candidate risks adapting the editor around a poor starting point. \textsc{Kinesys} therefore learns from $N$ candidate edits within each run. At the start of the run, it draws $N$ noise samples $\epsilon_i$ and reuses them throughout fine-tuning. At each step $t\in\{1,\ldots,T\}$, the current editor renders one candidate from each sample. Reusing the noise makes successive steps comparable and lets feedback track each candidate's progress as the editor changes.

\textbf{Reward-based candidate weighting.}
\label{sec:method_gating}
Within a run, all candidates shape the same LoRA. Equal weights would give an edit with artifacts or unintended source changes as much influence as a stronger edit. We smooth each EditReward score over optimization steps, standardize it across the $N$ candidates, and average the two scores per candidate. A softmax across these averages yields weights $\alpha_i$ that sum to one. This de-emphasizes poorly scored edits without discarding them; weights are recomputed between updates, then detached and held fixed during each LoRA update. Appendix~\ref{app:gate} gives the smoothing, normalization, and weighting equations.

\textbf{Maintaining candidate diversity.}
\label{sec:method_diversity}
An artist editing smoke around a car may want several valid results: each should reflect the reference smoke and maintain the original scene composition, while the smoke plume can vary in shape, density, and precise placement. The paired outputs in Figure~\ref{fig:fresh_noise_variations} illustrate variations from noise samples. During fine-tuning, however, shared objectives can draw all $N$ fixed-noise candidates toward similar appearances. \textsc{Kinesys} includes a diversity term $\mathcal{D}$ to discourage early convergence among the fixed-noise candidates.

Following repulsive particle objectives~\citep{liu2016steinvariational,corso2024particleguidance}, $\mathcal{D}$ averages radial basis function (RBF) similarities across distinct candidates. The complete optimization objective combines the weighted candidate losses with this diversity penalty:

\begin{subequations}
\noindent
\begin{minipage}{0.49\linewidth}
\begin{equation}
    \mathcal{L}_{\mathrm{pop}}
    = \sum_{i=1}^{N}\alpha_i\mathcal{L}_i
    + \lambda_{\mathrm{div}}(t)\mathcal{D}
    \label{eq:population_objective}
\end{equation}
\end{minipage}\hfill
\begin{minipage}{0.49\linewidth}
\begin{equation}
    \mathcal{D}
    = \frac{1}{N(N-1)}
      \sum_{i=1}^{N}\sum_{j\neq i}
      \exp\!\left(-\frac{d_{ij}}{h_i}\right)
    \label{eq:population_diversity}
\end{equation}
\end{minipage}
\end{subequations}

Here, $d_{ij}$ is the mean squared distance within the editable region between candidate $i$'s current latent and candidate $j$'s latent from the previous optimization step. The bandwidth $h_i$ is based on the median distance to the other candidates. We hold the previous-step latents and bandwidth fixed during gradient computation, so minimizing $\mathcal{D}$ pushes each current latent away from the other candidates' recent states. For $N=1$, we set $\mathcal{D}=0$. The gradient $\nabla_{\phi}\mathcal{L}_{\mathrm{pop}}$ updates only the LoRA parameters $\phi$. The weight $\lambda_{\mathrm{div}}(t)$ is positive during early updates to discourage candidate collapse, then decays to zero so later updates focus on reference-grounded VQA correctness and EditReward's instruction-following and visual-quality signals.

\textbf{Independent fine-tuning runs.}
\label{sec:method_search}
Candidate weighting reduces the influence of poor edits, but all $N$ candidates still shape one LoRA. Putting all samples into one run would keep their updates coupled, so a promising starting set could still be pulled toward a compromise with less useful ones. \textsc{Kinesys} instead starts $K=5$ independent runs, each with its own LoRA parameters $\phi_k$ and its own $N$ fixed-noise samples. These $K$ sets of $N$ noise draws allow different starting sets to improve independently; the runs exchange neither parameters nor candidates. The candidate and run study in Section~\ref{sec:experiments_ablations} supports this compute-limited configuration: three candidates score highest among the tested widths, and multiple independent runs improve the lower-of-two score over one run.

\textbf{Pruning and checkpoint selection.}
Not every starting set deserves the full tuning budget. \textsc{Kinesys} therefore saves intermediate LoRA checkpoints and uses the first two scheduled checkpoint boundaries to rank runs by recent VQA and EditReward feedback. Lower-ranked runs stop, leaving three after the first pruning point and two after the second. The counts limit compute, but the search remains open across multiple runs until a separate step selects one saved checkpoint.

To favor a checkpoint that works beyond the fixed noise samples used for fine-tuning, \textsc{Kinesys} renders an evaluation image using the same held-out noise sample at every saved checkpoint. It selects the LoRA parameters $\phi^{\star}$ whose evaluation image has the highest EditReward score. The selected parameters give the artist an adapted editor that can produce further effect variations from fresh noise while retaining the source scene. Appendices~\ref{app:method_details}, \ref{app:prompts}, and~\ref{app:full_qualitative} provide implementation details, prompt templates, and additional visual results.

\raggedbottom
\section{Experiments}
\label{sec:experiments}

We test whether \textsc{Kinesys} transfers a reference effect while preserving source content and structure. We compare it with open-weight and proprietary editors in Sec.~\ref{sec:experiments_main} and report ablations in Sec.~\ref{sec:experiments_ablations}.

\subsection{Experimental Setup}
\label{sec:experiments_setup}

\noindent\textbf{Dataset and tasks.}\enspace
\label{sec:experiments_data}
Our evaluation contains 81 effect-reference editing requests across eight categories: boat wakes, vehicle dents, explosions, fire, puddles, cast shadows, smoke, and snow. The requests use 78 distinct source photographs and 63 effect references. Each edit request includes a short effect-editing prompt and three yes/no questions reused within its effect category. These questions test placement, realism, and a defining effect property against the reference image. Appendix~\ref{app:dataset} provides more details on dataset construction.

\noindent\textbf{Baselines.}\enspace
\label{sec:experiments_baselines}
We compare \textsc{Kinesys} with six direct open-weight editors: FLUX.2 Klein 4B~\citep{flux-2-2025}, Qwen-Image-Edit~\citep{wu2025qwenimagetechnicalreport}, OmniGen2~\citep{wu2025omnigen2}, Step1X-Edit v1.2~\citep{liu2025step1xedit}, ACE++~\citep{mao2025aceplusplus}, and ICEdit~\citep{zhang2025icedit}, plus proprietary Nano Banana 2 (Gemini 3.1 Flash Image)~\citep{google2026nanobanana2}. All direct editors receive the source and instruction, and receive the effect reference where supported. They generate a result without per-request tuning or verification questions. Mask-capable editors receive a full-image mask; all baselines can edit the entire image.

Native FLUX.2 Klein 4B receives the same source, effect reference, mask, and generation settings as \textsc{Kinesys}, serving as its unadapted control. Our adapted Dual-Process baseline~\citep{luo2025dualprocessimagegeneration} fine-tunes the same image-conditioned generator with the same instruction, source, reference, and three questions as \textsc{Kinesys}, using only VQA feedback. For the automated comparison, \textsc{Kinesys} uses a full-image mask because some baselines do not support spatial masks.

\noindent\textbf{\textsc{Kinesys} configuration.}\enspace
For the main comparison and feedback ablation, a frozen Gemma 4 E4B-it VLM~\citep{gemmateam2026gemma4} answers the reference-grounded questions. Each run uses $N=3$ fixed-noise candidates for up to $T=100$ steps, saving checkpoints at the labels listed in Appendix~\ref{app:implementation}. \textsc{Kinesys} starts $K=5$ independent fine-tuning runs and uses VQA and EditReward feedback to retain three at the first pruning point and two at the second.

\begin{figure}[!t]
\makeatletter\def\@captype{table}\makeatother
\caption{Comparison on 156 shared cases from 81 requests and two seeds. Open-weight methods are ordered by Gemini; bold and underline mark first and second. $\dagger$: EditReward guides \textsc{Kinesys} tuning and selection (not independent). $\ddagger$: Nano Banana 2 and the judge share the Gemini family.}
\label{tab:main}
\vspace{1em}
\centering
\begingroup
\normalsize
\renewcommand{\arraystretch}{1.15}
\setlength{\tabcolsep}{2pt}
\begin{tabularx}{\columnwidth}{@{}>{\raggedright\arraybackslash}p{0.32\columnwidth}*{5}{>{\centering\arraybackslash}X}@{}}
\toprule
 & \multicolumn{3}{c}{Automated judges} & \multicolumn{2}{c}{Source preservation} \\
\cmidrule(lr){2-4}\cmidrule(lr){5-6}
Method & Gemini$\uparrow$ & \shortstack{Edit\\Score}$\uparrow$ & \shortstack{Edit\\Reward}$\uparrow$ & LPIPS$\downarrow$ & TPIPS$\uparrow$ \\
\midrule
\multicolumn{6}{@{}l}{\textit{Open-weight methods}} \\
OmniGen2 (2025) & $\phantom{-}4.95$ & $\phantom{-}7.53$ & $-0.018\phantom{{}^{\dagger}}$ & $\phantom{-}0.753$ & $\phantom{-}0.599$ \\
ACE++~\citeyearpar{mao2025aceplusplus} & $\phantom{-}4.98$ & $\phantom{-}5.95$ & $-0.503\phantom{{}^{\dagger}}$ & $\phantom{-}0.795$ & $\phantom{-}0.500$ \\
ICEdit~\citeyearpar{zhang2025icedit} & $\phantom{-}5.93$ & $\phantom{-}6.91$ & $-0.097\phantom{{}^{\dagger}}$ & $\phantom{-}\underline{0.429}$ & $\phantom{-}0.632$ \\
Dual-Process~\citeyearpar{luo2025dualprocessimagegeneration} & $\phantom{-}7.09$ & $\phantom{-}7.93$ & $\phantom{-}0.212\phantom{{}^{\dagger}}$ & $\phantom{-}0.577$ & $\phantom{-}0.615$ \\
Qwen-Image-Edit~\citeyearpar{wu2025qwenimagetechnicalreport} & $\phantom{-}7.67$ & $\phantom{-}8.31$ & $\phantom{-}0.435\phantom{{}^{\dagger}}$ & $\phantom{-}0.645$ & $\phantom{-}0.599$ \\
FLUX.2 Klein 4B (2026) & $\phantom{-}7.90$ & $\phantom{-}8.68$ & $\phantom{-}0.501\phantom{{}^{\dagger}}$ & $\phantom{-}0.579$ & $\phantom{-}0.640$ \\
Step1X-Edit~\citeyearpar{liu2025step1xedit} & $\phantom{-}8.22$ & $\phantom{-}8.80$ & $\phantom{-}0.666\phantom{{}^{\dagger}}$ & $\phantom{-}0.449$ & $\phantom{-}\underline{0.724}$ \\

\midrule
\multicolumn{6}{@{}l}{\textit{Proprietary baseline}} \\
Nano Banana 2~\citeyearpar{google2026nanobanana2} & $\phantom{-}\mathbf{8.91}{}^{\ddagger}$ & $\phantom{-}\mathbf{9.06}$ & $\phantom{-}\underline{0.738}\phantom{{}^{\dagger}}$ & $\phantom{-}\underline{0.429}$ & $\phantom{-}0.696$ \\
\midrule
\multicolumn{6}{@{}l}{\textit{Our method}} \\
\rowcolor{blue!5}
\textbf{\textsc{Kinesys} (Ours)} & $\phantom{-}\underline{8.83}$ & $\phantom{-}\underline{8.95}$ & $\phantom{-}\mathbf{1.007}{}^{\dagger}$ & $\phantom{-}\mathbf{0.377}$ & $\phantom{-}\mathbf{0.731}$ \\
\bottomrule
\end{tabularx}
\endgroup
\par\medskip
\includegraphics[width=\columnwidth]{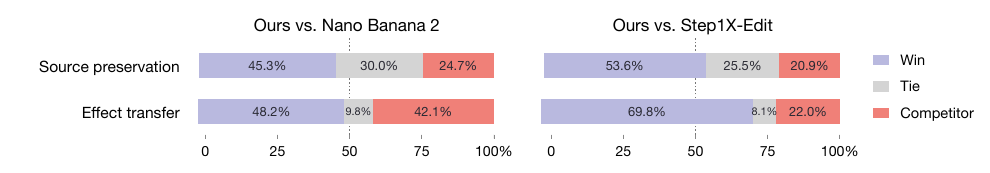}
\par\vspace{-0.6em}
\makeatletter\def\@captype{figure}\makeatother
\caption{Blinded paired-comparison user study on 16 edits, with 41 responses for Nano Banana 2 and 40 for Step1X-Edit. Bars show preferences for source preservation and effect transfer; ties indicate users voting for no clear winner.}
\label{fig:user_study_bars}
\vspace{-1em}
\end{figure}

\noindent\textbf{Evaluation protocol.}\enspace
\label{sec:experiments_metrics}
We run each of the 81 requests under two seeds to reduce dependence on a single random draw. For each seed, \textsc{Kinesys} starts a separate set of fine-tuning runs and selects a checkpoint independently. We compare methods on the 156 of 162 cases completed by every method, excluding cases with missing or unusable provider responses. Each reported score is the arithmetic mean over the retained results from each request and seed, with \mbox{equal weight for each result.}

We prompt a held-out Gemini 2.5 Flash judge~\citep{googledeepmind2025gemini25} with the source image, effect reference, edited image, and instruction. The judge reports scores from 0 to 10 for perceptual quality (realism and artifacts) and source consistency (preservation of untargeted content). Table~\ref{tab:main} reports the average of these two scores.
EditScore-72B~\citep{luo2026editscoreunlockingonlinerl} and EditReward~\citep{wu2026editrewardhumanalignedrewardmodel} are learned judges that receive the source, instruction, and edited image; neither sees the effect reference. EditScore-72B combines semantic consistency (instruction following and preservation of unrequested content) with perceptual quality (realism and artifacts) into an overall edit-quality score. EditReward is trained on human preferences and predicts separate instruction-following and visual-quality scores, which we average.

We compare each edit with its source using LPIPS distance~\citep{zhang2018unreasonable} and TPIPS under its ``overall'' setting~\citep{wang2026tpips}. LPIPS measures distance in learned visual features, while TPIPS measures perceptual similarity conditioned on a text prompt. We use both as proxies for overall source preservation; neither score directly measures camera pose, object layout, or subject identity. Gemini and EditScore-72B judge broader editing quality that includes source consistency. An unchanged edit can score well on similarity while failing to transfer the effect, so we interpret these measures alongside the visual results and user study.

\noindent\textbf{User study.}\enspace
Because automated judges are imperfect proxies for human judgment, we also ran a blinded user study; the final analysis includes 16 edits, two per effect category. We compared \textsc{Kinesys} with Step1X-Edit v1.2, the highest-scoring evaluated open-weight editor in Table~\ref{tab:main}, and with proprietary Nano Banana 2. We report 41 completed response sets for Nano Banana 2 and 40 for Step1X-Edit; most respondents are not VFX experts. Each response set contributes 16 paired comparisons, with separate source-preservation and effect-transfer judgments and ties allowed. Figure~\ref{fig:user_study_bars} reports preferences. Appendix~\ref{app:user_study} gives the survey protocol, preferences by effect family, and an example survey question.

\begin{figure}[!t]
    \centering
    \includegraphics[width=\columnwidth]{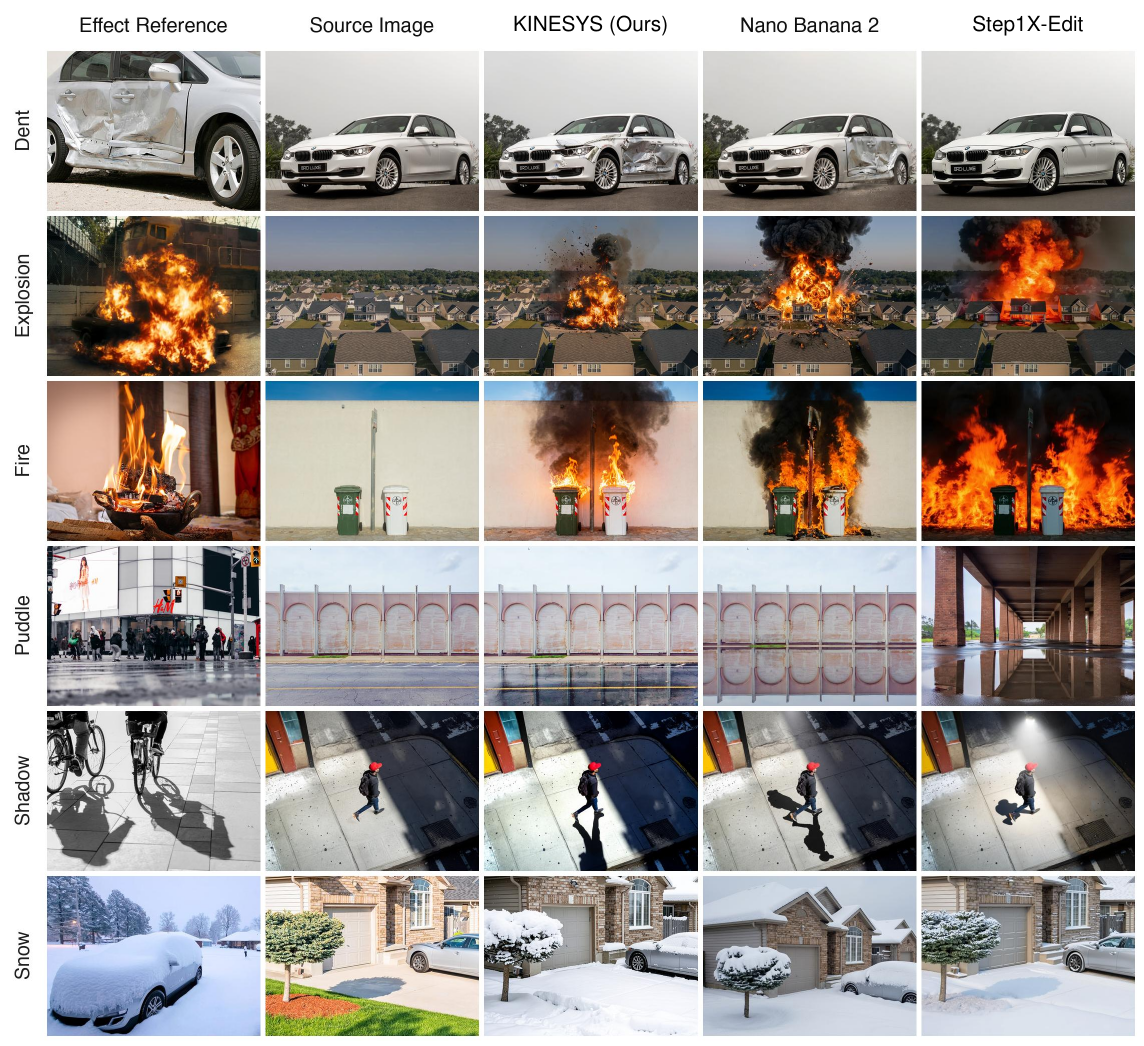}
    \par\vspace{-0.5em}
    \caption{Six-effect comparison. Columns: reference, source, \textsc{Kinesys}, Nano Banana 2, Step1X-Edit v1.2. For snow, \textsc{Kinesys} retains the source house and viewpoint while Nano Banana 2 changes both. For fire, \textsc{Kinesys} adds flames to trash cans without Step1X-Edit's scene-wide blaze.}
    \label{fig:qualitative_main}
\end{figure}

\subsection{Comparison with Existing Editors}
\label{sec:experiments_main}

{\widowpenalties 3 10000 10000 0
\textsc{Kinesys} achieves the strongest source preservation scores among the evaluated editors: LPIPS distance is $0.377$ versus Nano Banana 2's $0.429$, and TPIPS similarity is $0.731$ versus $0.696$. It also leads the evaluated open-weight editors on Gemini and EditScore-72B. The user study in Figure~\ref{fig:user_study_bars} reinforces the preservation result. \textsc{Kinesys} was preferred for source preservation over Nano Banana 2 ($45\%$ versus $25\%$) and Step1X-Edit ($54\%$ versus $21\%$). For effect transfer, preferences were close against Nano Banana 2 ($48\%$ versus $42\%$) and favored \textsc{Kinesys} over Step1X-Edit ($70\%$ versus $22\%$). Figure~\ref{fig:main_teaser} also shows scene changes in the `puddle' edit and an extra boat in Nano Banana 2's `boat-wake' edit illustrating these differences in source preservation.\par}

\begin{figure}[t!]
    \centering
    \includegraphics[width=\columnwidth,trim=120bp 120bp 120bp 110bp,clip]{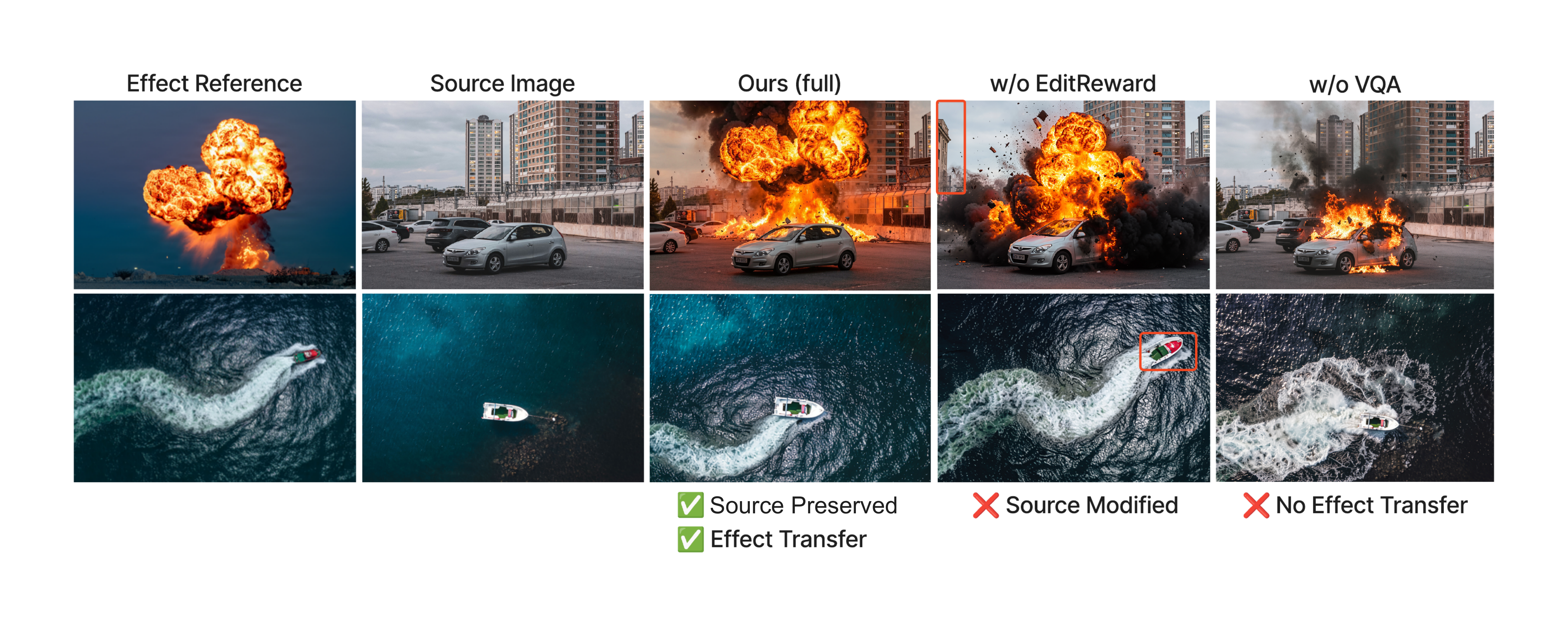}
    \caption{Feedback ablations on explosion and wake edits. The ``w/o'' labels denote removal of the named gradient; red boxes mark unintended source changes without EditReward. Without VQA, effect form departs from the reference. Full \textsc{Kinesys} preserves source content and transfers the effect.\looseness=-1}
    \label{fig:feedback_ablations}
\end{figure}

\subsection{Ablation}
\label{sec:experiments_ablations}

\noindent\textbf{Effect transfer and source preservation.}\enspace We highlight the need for two complementary VLM judges in Figure~\ref{fig:feedback_ablations}. Without the EditReward gradient, the effect appears but unrelated source content changes: a building appears in the explosion edit, and the boat changes appearance and position in the wake edit. Without the VQA gradient, the scene stays closer to the source, but the explosion loses the reference's form, and the wake misses its intended pattern. With both gradients, \textsc{Kinesys} better balances effect transfer and source preservation in these examples.

\begin{table}[!t]
    \centering
    \caption{Gemini scores for 16 requests, varying candidate count $N$ and independent runs $K$. Mean averages both seeds; Worst averages the lower seed score per request.}
    \label{tab:search_configuration}
    \renewcommand{\arraystretch}{1.15}
    \begin{minipage}[t]{0.46\columnwidth}
        \centering
        \textbf{(a) Candidate edits per run}\\[5pt]
\begin{tabular*}{\linewidth}{@{\extracolsep{\fill}}l c c@{}}
            \toprule
            $N$ & Mean$\uparrow$ & Worst$\uparrow$ \\
            \midrule
            $1$ & $8.59$ & $8.11$ \\
            \rowcolor{blue!5}
            \textbf{$3$ (ours)} & $9.00$ & $8.73$ \\
            $5$ & $8.55$ & $7.80$ \\
            \bottomrule
        \end{tabular*}
    \end{minipage}\hfill
    \begin{minipage}[t]{0.46\columnwidth}
        \centering
        \textbf{(b) Independent fine-tuning runs}\\[5pt]
\begin{tabular*}{\linewidth}{@{\extracolsep{\fill}}l c c@{}}
            \toprule
            $K$ & Mean$\uparrow$ & Worst$\uparrow$ \\
            \midrule
            $1$ & $7.31$ & $6.45$ \\
            $3$ & $9.02$ & $8.59$ \\
            \rowcolor{blue!5}
            \textbf{$5$ (ours)} & $9.00$ & $8.73$ \\
            $8$ & $9.12$ & $8.88$ \\
            \bottomrule
        \end{tabular*}
    \end{minipage}
\end{table}

\begin{figure}[!t]
    \centering
    \includegraphics[width=\columnwidth]{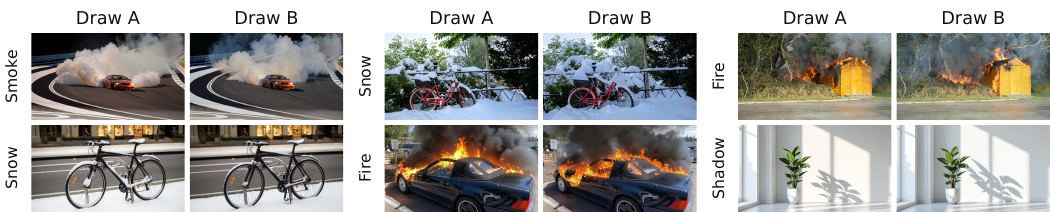}
    \par\vspace{-1em}
    \caption{Fresh-noise variations in smoke shape, snow intensity, fire extent, and cast shadows. Each pair uses the same source, reference, instruction, and selected LoRA; only the noise changes.}
    \label{fig:fresh_noise_variations}
    \vspace{-1em}
\end{figure}

\noindent\textbf{Candidate and fine-tuning run configurations.}\enspace The numbers of candidates and independent runs affect both editing quality and per-request compute. We use $N=3$ candidates and $K=5$ runs, and vary each setting separately in Table~\ref{tab:search_configuration} to assess that balance. Increasing $N$ from 1 to 3 raises mean Gemini from $8.59$ to $9.00$ and the mean worst-seed score from $8.11$ to $8.73$. At $N=5$, these fall to $8.55$ and $7.80$. One possible explanation is that feedback from a wider set of fixed-noise candidates pulls their shared adaptation toward a less useful compromise.

Increasing $K$ from one to five runs raises mean Gemini from $7.31$ to $9.00$ and the mean worst-seed score from $6.45$ to $8.73$. Three runs have a slightly higher mean ($9.02$) but a lower worst-seed score ($8.59$); eight runs improve both to $9.12$ and $8.88$ at greater compute. We use five runs to favor worst-seed quality without the cost of eight. Appendix~\ref{app:search_configuration} gives the seeds and pruning schedules.

\vspace{-0.8em}
\subsection{Applications}
\label{sec:applications}

\begin{figure}[H]
    \centering
    \includegraphics[width=\columnwidth]{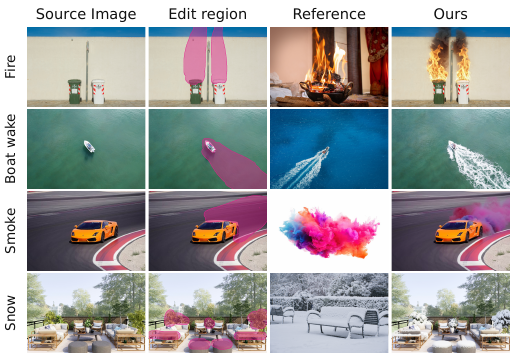}
    \caption{Spatially controlled effect-reference edits. Columns show the source, pink edit region, effect reference, and \textsc{Kinesys} result.}
    \label{fig:masked_edits}
\end{figure}

Visual-effects artists may know the desired effect and where it belongs, yet want several convincing versions before choosing one (Figure~\ref{fig:main_teaser}, top). \textsc{Kinesys} offers both forms of control: a mask localizes the effect while the surrounding scene remains consistent (Figure~\ref{fig:masked_edits}), and sampling the selected checkpoint with fresh noise changes details such as snow amount, smoke shape and density, or burn area (Figure~\ref{fig:fresh_noise_variations}). The requested effect and source scene remain consistent across those draws. The fresh-noise evaluation uses all 162 selected \textsc{Kinesys} checkpoints, one per request and seed. Editing quality holds across five unseen draws per checkpoint: mean EditScore-72B is $8.91$, compared with $8.93$ for the selected outputs.

\clearpage

\vspace{-0.8em}
\section{Conclusion}
\label{sec:conclusion}

\textsc{Kinesys} adapts a generator to transfer a reference-demonstrated effect while preserving the source shot's composition and content. Complementary VLM feedback and independent fine-tuning runs support this balance, while fresh-noise samples offer multiple edit variations. On the shared evaluation cases, \textsc{Kinesys} achieves better LPIPS and TPIPS scores than Nano Banana 2 and is preferred for source preservation in the blinded user study, while remaining competitive on editing quality and effect transfer. It leads the evaluated open-weight editors on Gemini and EditScore-72B.

The approach still depends on what its generator can render and its VLM judges can recognize: an unfamiliar creature with a rare stylized electrical effect, for example, may be difficult to edit faithfully. Bad starting noise can still yield a weak edit, and per-request fine-tuning adds latency. \textsc{Kinesys} is a first step towards frameworks that enable effect transfer while preserving \mbox{source content and structure.}

\section{Acknowledgments}
\label{sec:acknowledgments}

This work was partly supported by NCInnovation under the project ``KINESYS+ATLAS: AI and Synthetic Data for Critical Applications''.

\begingroup
\addtolength{\bibsep}{-1.5pt}
\bibliography{iclr2027_conference}
\bibliographystyle{iclr2027_conference}
\endgroup

\clearpage
\appendix
\numberwithin{equation}{section}
\numberwithin{figure}{section}
\numberwithin{table}{section}
\raggedbottom
\section*{Appendix}

Section~\ref{app:full_qualitative} expands the visual comparison with all methods shown side by side. Section~\ref{app:dataset} describes how the 81 requests were paired and displays their distribution. Section~\ref{app:method_details} gives the gradient boundary, weighting, search, and implementation settings behind the method. Section~\ref{app:prompts} describes generation inputs and records the model-facing optimization and evaluation prompts. Section~\ref{app:user_study} documents the blinded survey and shows preferences by effect family.

\section{Additional Visual Results}
\label{app:full_qualitative}

\begin{figure}[H]
    \centering
    \includegraphics[width=\textwidth,trim=0 794.64bp 0 0,clip]{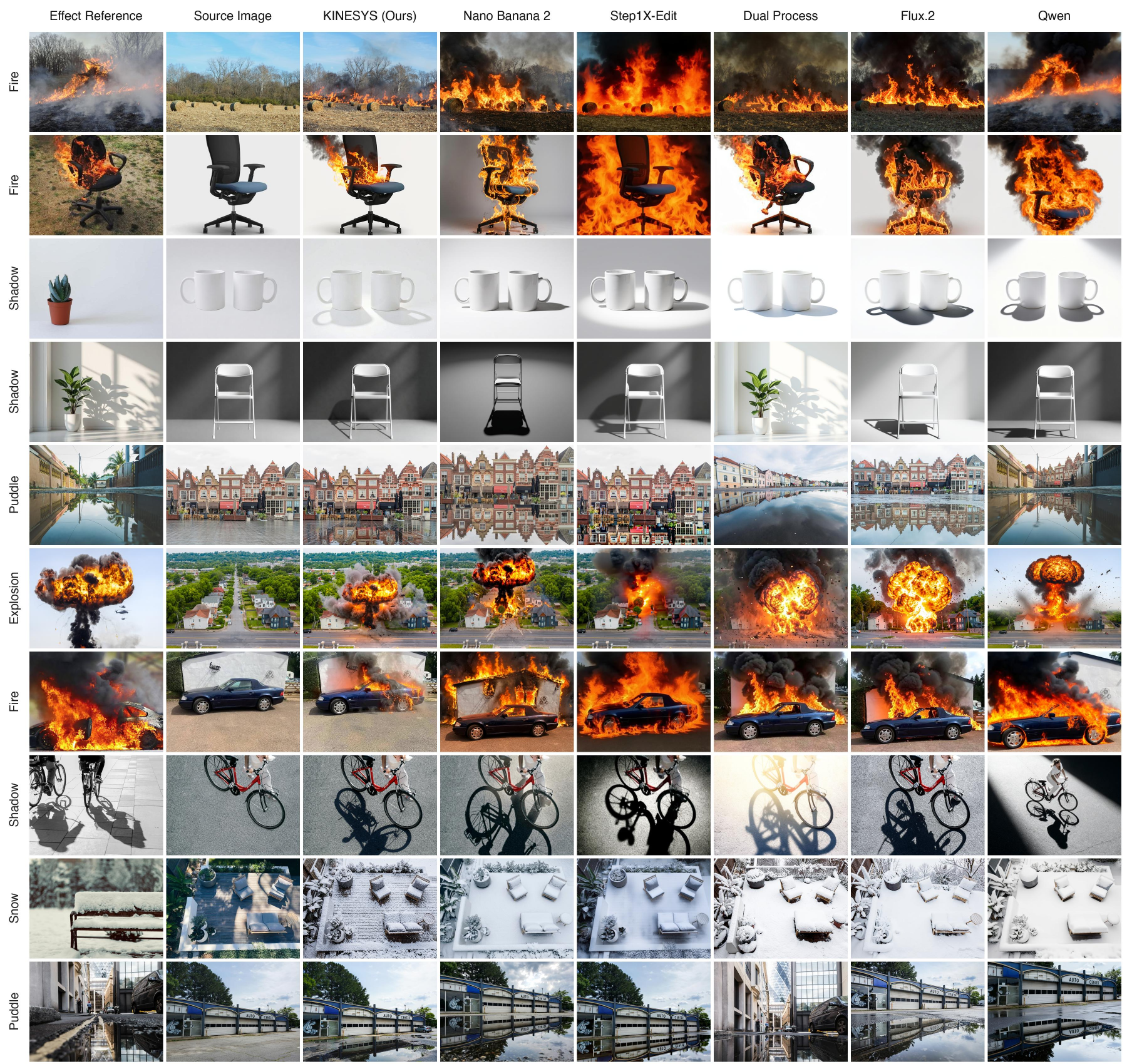}
    \caption{Qualitative comparisons, rows 1 to 10. Columns: effect reference, source, \textsc{Kinesys}, Nano Banana 2, Step1X-Edit v1.2, Dual-Process, FLUX.2 Klein 4B, and Qwen-Image-Edit.}
    \label{fig:qualitative_full}
\end{figure}

\begin{figure}[H]
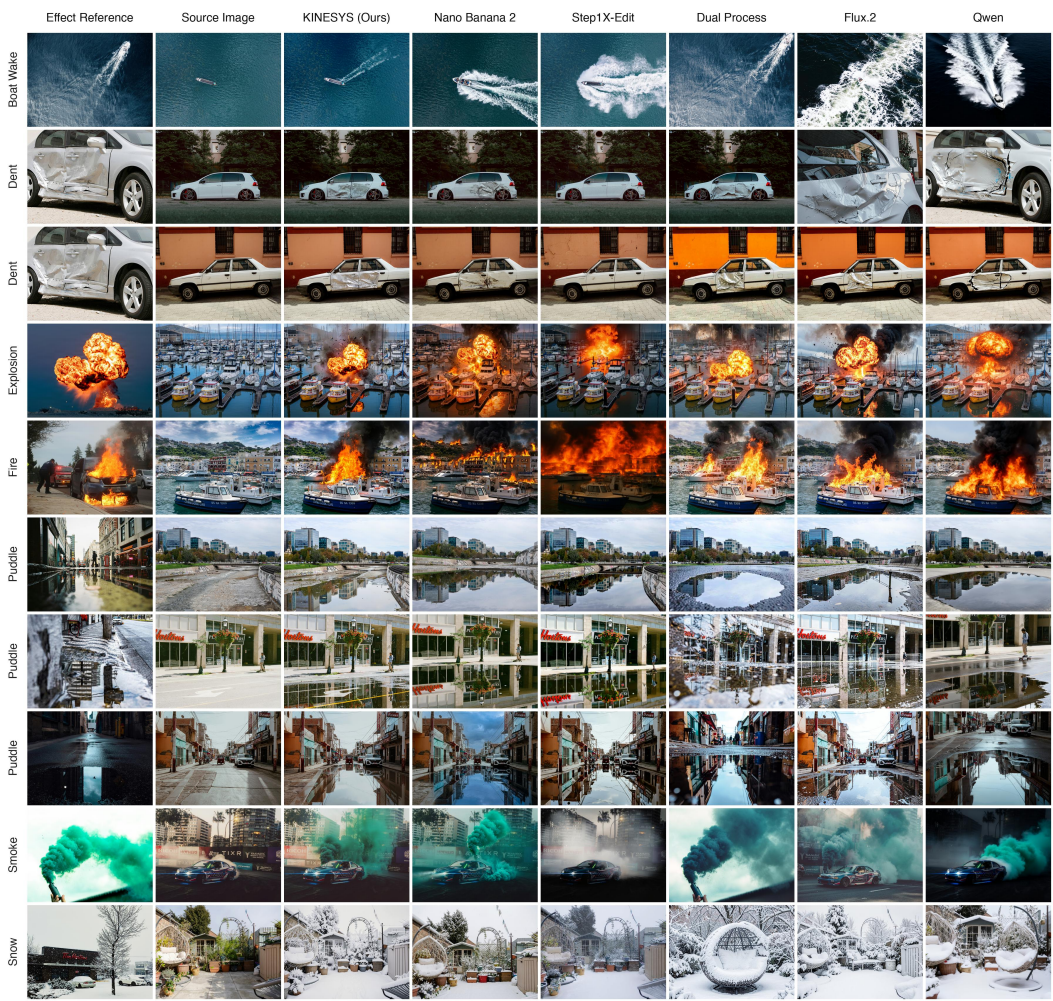

    \centering
    \leavevmode\vbox{%
      \hbox{\includegraphics[width=\textwidth,trim=0 1586.64bp 0 0,clip]{figures/large_grid.pdf}}%
      \nointerlineskip
      \hbox{\includegraphics[width=\textwidth,trim=0 2.64bp 0 816bp,clip]{figures/large_grid.pdf}}%
    }
    \caption{Qualitative comparisons, rows 11 to 20. Columns: effect reference, source, \textsc{Kinesys}, Nano Banana 2, Step1X-Edit v1.2, Dual-Process, FLUX.2 Klein 4B, and Qwen-Image-Edit.}
    \label{fig:qualitative_full_b}
\end{figure}

Figures~\ref{fig:qualitative_full} and~\ref{fig:qualitative_full_b} show 20 effect-reference comparisons. Each row keeps the same source and effect reference across methods. Both figures show all eight columns in the same order, with the grid split only between rows.

\clearpage

\clearpage
\section{Evaluation Dataset Construction}
\label{app:dataset}

\begin{figure}[H]
    \centering
    \includegraphics[width=0.76\textwidth]{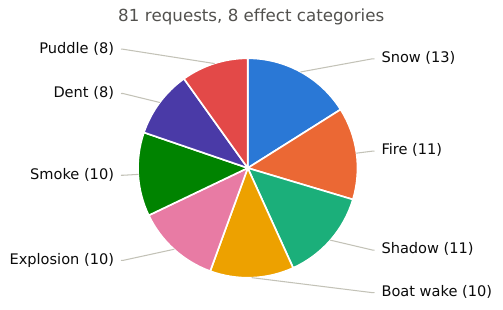}
    \caption{The 81 requests by effect family. Counts are requests, not unique photographs.}
    \label{fig:dataset_pie}
\end{figure}

\noindent\textbf{Requests and effect families.}\enspace
We evaluate 81 requests across eight effect families. Each pairs a source photograph with a reference photograph showing the desired effect on different content; there is no target edited image. These requests use 78 distinct source photographs because three car photographs appear in both the dent and fire families. They use 63 distinct effect references; 13 of those references serve between two and four requests within their family. Figure~\ref{fig:dataset_pie} shows the request split, and Table~\ref{tab:dataset_families} gives the family counts and the properties selected for verification. Family sizes range from eight to 13 requests; reported automated scores weight the retained results from each request and seed equally.

\noindent\textbf{Source and reference pairing.}\enspace
We selected each request's effect reference by reviewing the source photograph alongside references from the same effect family, without using a generated target edit. A reference may serve multiple requests within its family.

\noindent\textbf{Instructions and questions.}\enspace
We used one instruction template per effect family. It describes the requested effect, treats the reference as an exemplar of selected effect properties, and avoids describing the reference's own scene. Editors and judges received the same instruction.

In the intended workflow, users define questions to select which properties of the reference effect to transfer. For this benchmark, we authored three fixed yes/no questions per family. They check where the effect appears, whether it looks realistic, and whether a defining property matches the reference. For example, the questions about cast shadows ask whether the shadow falls beneath the subject, is sharp and well-defined, and matches the reference's direction and length. The VLM receives the reference as Image~1 and the candidate edit as Image~2; the desired answer to each question is ``Yes.'' The questions are reused across requests within a family and remain fixed during fine-tuning. Section~\ref{app:prompts} gives the model-facing question format.

\noindent\textbf{Image preparation and mask policy.}\enspace
Source and reference photographs were downscaled, without upscaling, to a longest side of at most 768 pixels. The editor's per-request image size was then rounded down to a multiple of 16. The adapted editor used a reference-conditioning longest-side cap of 384 pixels, while the in-loop VLM compared the candidate with the reference at its working resolution of at most 768 pixels. The automated benchmark specifies a whole-image edit region; mask-capable editors receive a full-image mask, including for effects that could be localized. The hand-drawn masks in the Applications figure are examples outside these 81 requests.

\clearpage

\begin{table}[H]
\caption{Composition of the 81 requests. Counts are requests and distinct reference images. The listed properties summarize location and questions specific to each effect.}
\label{tab:dataset_families}
\centering
\normalsize
\renewcommand{\arraystretch}{1.15}
\setlength{\tabcolsep}{4pt}
\begin{tabularx}{\columnwidth}{@{}lcc>{\raggedright\arraybackslash}X@{}}
\toprule
Effect family & Requests & References & Family-specific checks \\
\midrule
Boat wake & 10 & 10 & Placement behind the boat; turbulence and spread \\
Vehicle dent & 8 & 4 & Body-panel location; size~and~depth \\
Explosion & 10 & 8 & Location; scale and intensity \\
Fire & 11 & 8 & Flames on the subject; size and intensity \\
Puddle & 8 & 7 & Foreground placement; reflection clarity \\
Cast shadow & 11 & 8 & Position beneath the subject; sharpness, direction,~and~length \\
Smoke & 10 & 8 & Origin at the subject; color and density \\
Snow & 13 & 10 & Surface coverage; depth and texture \\
\midrule
Total & 81 & 63 & \\
\bottomrule
\end{tabularx}
\end{table}

\clearpage
\section{Optimization and Implementation Details}
\label{app:method_details}

\subsection{Diversity Term and Decay}
\label{app:diversity}

Let $z_i^{(t)}$ be candidate $i$'s final estimate of the clean latent at step $t$.
Each fine-tuning run maintains its own bank of the previous step's $N$
candidate latents. The bank is replaced after each step and is not shared
between runs. It starts empty, so the diversity penalty is zero on the
first update. Define the detached latents from the previous step as
$\bar z_j^{(t-1)}:=\operatorname{sg}(z_j^{(t-1)})$.
Distances are computed only inside the downsampled edit mask:
\begin{equation}
    d_{ij}^{(t)} = \operatorname{MSE}_{m}\!\left(z_i^{(t)},\bar z_j^{(t-1)}\right),
    \qquad
    \mathcal{D}_i^{(t)} = \frac{1}{N-1}\sum_{j\neq i}
    \exp\!\left(-\frac{d_{ij}^{(t)}}{h_i^{(t)}}\right).
    \label{eq:diversity}
\end{equation}
The population penalty in the main paper is
$\mathcal{D}^{(t)}:=\frac{1}{N}\sum_{i=1}^{N}\mathcal{D}_i^{(t)}$. At $N=1$,
there are no candidate pairs, so we do not compute pairwise distances or a
bandwidth and set $\mathcal{D}^{(t)}=0$. The stop-gradient operator $\operatorname{sg}(x)$ leaves $x$ unchanged in the forward pass and gives it zero derivative during backpropagation. A latent from the previous step saved without its computation graph is already detached in practice; the notation states that gradient boundary explicitly. Each diversity term has a gradient through its current candidate latent, but not through the bank of latents from the previous step. For each candidate, the RBF bandwidth follows the median heuristic:
\begin{equation}
    h_i^{(t)} = \operatorname{sg}\!\left[\max\!\left(
      \operatorname*{median}_{j\neq i} d_{ij}^{(t)},\; h_{\min}
    \right)\right],
    \label{eq:bandwidth}
\end{equation}
and is treated as constant during backward propagation~\citep{liu2016steinvariational}.
We use $h_{\min}=10^{-4}$.
The current latent therefore receives gradients through the distances in the
RBF numerator, while neither the bank of latents from the previous step nor the bandwidth receives a
gradient.

When the median remains above $h_{\min}$ before and after a common rescaling of
all distances, the \emph{forward value} of the RBF term is unchanged.
The floor prevents ill-conditioned normalization when candidates become very
similar; as with any squared-distance RBF, the repulsive gradient can still
vanish at exact coincidence. We initialize $\lambda_{\mathrm{div}}^{0}=1$ and
decay the diversity weight during the first
$60\%$ of optimization,
\begin{equation}
    \lambda_{\mathrm{div}}(t)=\lambda_{\mathrm{div}}^{0}
    \max\!\left(0,1-\frac{t}{0.6T}\right),
    \label{eq:diversity_decay}
\end{equation}
so this term contributes only during early optimization steps.

\subsection{Candidate Gate}
\label{app:gate}

The main paper summarizes the candidate gate. The EditReward component scores $R_{i,c}^{(t)}$ aggregate to
$R_{\mathrm{ER},i}^{(t)}$, with higher values indicating better edits.
Each fine-tuning run maintains a separate exponential moving average for
each candidate and reward head. We initialize each average with its first
observed reward, without bias correction. After initialization, we update
these averages using
\begin{equation}
    \bar{R}_{i,c}^{(t)} = \beta\,\bar{R}_{i,c}^{(t-1)}
    + (1-\beta)\,R_{i,c}^{(t)},
    \label{eq:ema}
\end{equation}
then standardize them within the current candidate population and fuse them across heads,
\begin{equation}
    A_i^{(t)} =
    \sum_c w_c\,
    \frac{\bar{R}_{i,c}^{(t)}-\mu_{\bar{R},c}^{(t)}}
         {\max\!\left(\sigma_{\bar{R},c}^{(t)},\,\sigma_{0,c}\right)},
    \qquad
    \alpha_i^{(t)} =
    \frac{\exp\!\left(A_i^{(t)}/\tau\right)}
         {\sum_{j=1}^{N}\exp\!\left(A_j^{(t)}/\tau\right)},
    \label{eq:gate}
\end{equation}
where $\mu_{\bar{R},c}^{(t)}$ and $\sigma_{\bar{R},c}^{(t)}$ are the mean and population standard deviation of $\{\bar{R}_{i,c}^{(t)}\}_{i=1}^{N}$, with variance normalized by $N$. The noise floor $\sigma_{0,c}$
bounds how sharply the gate can respond: when the spread of EMA rewards across
candidates falls below it, the standardized advantages shrink and $\alpha^{(t)}$
approaches a uniform distribution, giving equal weight to candidate task
gradients. The diversity gradient remains separate.
We use $\beta=0.9$, $\tau=1$, $\sigma_{0,c}=0.15$, and equal head weights
$w_{\mathrm{follow}}=w_{\mathrm{quality}}=0.5$.
Before each LoRA update, the gate weights are detached and held fixed. Gradients
flow through candidate losses but not through the EMA or gate.

\subsection{Fine-tuning Run Allocation and Checkpoint Selection}
\label{app:allocator}

At checkpoint $t$, fine-tuning run $k$ is summarized over the trailing window of
$W=15$ optimization steps. For each question, we average the VLM's desired-answer
probability over all $N$ candidates and the $W$ steps, then average across all
$Q$ questions. We average the higher-is-better EditReward score $R_{\mathrm{ER}}$
over the same candidates and steps:
\begin{equation}
    \bar{v}_k^{(t)} = \frac{1}{Q}\sum_{\ell=1}^{Q}
      \frac{1}{W}\sum_{\tau=t-W+1}^{t} v_{k,\ell}^{(\tau)},
    \qquad
    \bar{R}_k^{(t)} = \frac{1}{W}\sum_{\tau=t-W+1}^{t}
      R_{\mathrm{ER},k}^{(\tau)},
    \label{eq:window_stats}
\end{equation}
where the step-level summaries explicitly average over the $N$ candidates,
\begin{equation}
    v_{k,\ell}^{(\tau)} = \frac{1}{N}\sum_{i=1}^{N}
      \exp\!\left(-\mathcal{L}_{\mathrm{VQA},k,i,\ell}^{(\tau)}\right),
    \qquad
    R_{\mathrm{ER},k}^{(\tau)} = \frac{1}{N}\sum_{i=1}^{N}
      R_{\mathrm{ER},k,i}^{(\tau)}.
    \label{eq:step_candidate_averages}
\end{equation}
Here $\mathcal{L}_{\mathrm{VQA},k,i,\ell}^{(\tau)}$ is the negative log-likelihood
for question $\ell$ in run $k$ and candidate $i$, as in
the main-paper VQA objective; its exponential therefore gives the desired-answer
probability used for allocation.
We standardize and blend both summaries across the active fine-tuning runs
$\mathcal{K}^{(t)}$:
\begin{equation}
    s_k^{(t)} =
    w_{\mathrm{VQA}}
    \frac{\bar{v}_k^{(t)}-\mu(\bar{v}^{(t)})}
         {\sigma(\bar{v}^{(t)})}
    + w_{\mathrm{ER}}
    \frac{\bar{R}_k^{(t)}-\mu(\bar{R}^{(t)})}
         {\sigma(\bar{R}^{(t)})},
    \quad \mu,\sigma \text{ over } k\in\mathcal{K}^{(t)}.
    \label{eq:allocator}
\end{equation}
Here $\mu$ and $\sigma$ are the mean and population standard deviation over
the active runs. If either standard deviation is zero, its standardized term
is defined as zero for every run. A constant metric therefore does not affect
the ranking; with one active run, both standardized terms are zero.
We use $w_{\mathrm{VQA}}=w_{\mathrm{ER}}=\tfrac12$. Because the population
changes after pruning, this score is a within-checkpoint rank and is not
compared numerically across checkpoints. The search starts with five fine-tuning runs and retains three and then two
at zero-indexed checkpoint labels 15 and 30, respectively. Because the implementation begins with the previous boundary
set to $-1$, label 15 follows 16 parameter updates; we preserve these
historical labels in the reported protocol.

At each checkpoint boundary, the LoRA parameters of every surviving fine-tuning run
are saved and become eligible for final checkpoint selection. The full pipeline
reuses one dedicated noise sample, separate from the $N$ samples used during
fine-tuning, to render an evaluation image for every saved checkpoint. This image
contributes
neither gradients nor allocation feedback.
Its EditReward is recorded in \texttt{eval\_scores.json}. Let
$\mathcal{S}_{\mathrm{ckpt}}$ be the set of saved parameter checkpoints with an
available EditReward score, and let $y_{\mathrm{eval}}(\phi)$ denote the
fixed-input evaluation image \mbox{for checkpoint $\phi$. We select}
\begin{equation}
    \phi^{\star}
    = \operatorname*{arg\,max}_{\phi\in\mathcal{S}_{\mathrm{ckpt}}}
      R_{\mathrm{ER}}\!\left(y_{\mathrm{eval}}(\phi)\right).
    \label{eq:checkpoint_selection}
\end{equation}
A saved checkpoint without an EditReward entry in
\texttt{eval\_scores.json} is excluded; other scored checkpoints from the same
fine-tuning run remain eligible. VQA is used by the within-checkpoint allocator in
Equation~\ref{eq:allocator}, but not by this cross-checkpoint selector. Held-out
evaluation models enter neither allocation nor final checkpoint selection.

This noise sample is held out from fine-tuning but is used for checkpoint
selection. The fresh-noise evaluation in Section~\ref{sec:applications}
uses five additional unseen samples for each selected checkpoint.

Each surviving run may terminate before the nominal budget of 100 updates using a
rule based on reward drawdown. The monitored reward is the mean EditReward over the
run's $N$ candidates, smoothed by a mean over the trailing 10 steps. Stopping cannot
occur before step 50 and requires this smoothed value to remain more than
$0.3$ below its running peak for three consecutive steps. On stopping, the
run restores its best saved LoRA state from a chunk boundary.

\clearpage
\subsection{Implementation Details}
\label{app:implementation}

\noindent\textbf{Generator and optimization.}\enspace
The generator is FLUX.2 Klein 4B~\citep{flux-2-2025}, using checkpoint \nolinkurl{black-forest-labs/FLUX.2-klein-4B}. It runs in bfloat16 with four flow steps and guidance scale 1.0. Inpainting runs at the source resolution with the reference-conditioning longest side capped at 384 pixels. The LoRA~\citep{hu2021loralowrankadaptationlarge} has rank 16, is applied to transformer attention and feed-forward projections, and uses AdamW~\citep{loshchilov2019decoupled} with learning rate $5\times10^{-5}$. Each fine-tuning run uses up to $T=100$ optimization steps, with checkpoints at zero-indexed labels 15, 30, 45, 60, 75, 90, and 99, corresponding to 16, 31, 46, 61, 76, 91, and 100 updates. Only the LoRA matrices are updated; the base generator and text encoder remain frozen. The Dual-Process baseline uses the same image-conditioned generator and inputs but only VQA feedback; \textsc{Kinesys} also uses EditReward feedback and the candidate and run search.

\noindent\textbf{Feedback models and search schedule.}\enspace
The in-loop VLM is frozen Gemma 4 E4B-it~\citep{gemmateam2026gemma4}, using checkpoint \nolinkurl{google/gemma-4-E4B-it}. For each question, it receives the effect reference as Image~1 and the candidate edit as Image~2, then answers ``Yes'' or ``No.'' Differentiable EditReward uses the released Qwen2.5-VL-7B checkpoint with equal instruction-following and quality weights, $w_{\mathrm{follow}}=w_{\mathrm{quality}}=0.5$~\citep{wu2026editrewardhumanalignedrewardmodel}. Each fine-tuning run uses $N=3$ fixed-noise candidate edits. The task objective uses $\lambda_{\mathrm{VQA}}=1$ and $\lambda_{\mathrm{ER}}=2$, while the diversity weight begins at 1 and decays to zero after $60\%$ of the fine-tuning schedule. Search begins with five fine-tuning runs and prunes $5\!\rightarrow\!3$ at checkpoint label 15 and $3\!\rightarrow\!2$ at label 30.

\subsection{Search Configuration Study}
\label{app:search_configuration}

For Table~\ref{tab:search_configuration}, we fix $K=5$ when varying $N$
and fix $N=3$ when varying $K$; other control settings are unchanged.
Each request uses seeds 42 and 3047. The reported analysis covers the 16 requests
in Table~\ref{tab:search_requests}, two per effect family, from a method-blind
draw with sampling seed 20260830. This gives 32 results~across~the~two~seeds.

\begin{table}[H]
\centering
\caption{Requests in the search configuration analysis, each evaluated under both seeds.}
\label{tab:search_requests}
\begingroup
\renewcommand{\arraystretch}{1.15}
\setlength{\tabcolsep}{10pt}
\begin{tabular}{@{}lll@{}}
\toprule
Effect family & Request 1 & Request 2 \\
\midrule
Boat wake & \texttt{boatwake.4} & \texttt{boatwake.10} \\
Vehicle dent & \texttt{dent.7} & \texttt{dent.8} \\
Explosion & \texttt{explosion.3} & \texttt{explosion.9} \\
Fire & \texttt{fire.8} & \texttt{fire.9} \\
Puddle & \texttt{puddle.2} & \texttt{puddle.5} \\
Cast shadow & \texttt{shadow.5} & \texttt{shadow.11} \\
Smoke & \texttt{smoke.5} & \texttt{smoke.9} \\
Snow & \texttt{snow.6} & \texttt{snow.8} \\
\bottomrule
\end{tabular}
\endgroup
\end{table}

The allocation schedule uses checkpoint labels 15 and 30 for every tested $K$,
retaining $\lceil K/2\rceil$ and $\min(2,\lceil K/2\rceil)$ runs,
respectively. Thus, the retained counts are $(1,1)$ for $K=1$,
$(2,2)$ for $K=3$, $(3,2)$ for $K=5$, and $(4,2)$ for $K=8$.

\clearpage
\section{Optimization and Evaluation Prompts}
\label{app:prompts}

This section records the model-facing evaluation and optimization templates used in the canonical experiments. Angle-bracketed fields are replaced at runtime. Line wrapping,
markup, and punctuation have been normalized for typesetting; the substantive
wording, scoring criteria, and image order are retained. The VQA and EditReward prompts affect fine-tuning gradients, and EditReward also selects the final checkpoint; the
Gemini prompts are evaluation-only and never enter fine-tuning, allocation, or
final checkpoint selection.

\subsection{Generation Inputs}

Each editor receives the request's source, effect reference where supported, and the same fixed instruction; mask-capable editors also receive the full-image mask. The adapted FLUX.2 Klein 4B generator receives both images as inputs.

\subsection{Reference-Grounded VQA}

For every binary criterion, the frozen VLM receives the effect reference first
and the current candidate second. The VQA objective averages the negative
log-likelihood of the desired answers across the three fixed questions for
each request.
The model-specific chat template wraps the following user content. The assistant turn contains the target answer; cross-entropy uses only its answer tokens.

\begin{lstlisting}[style=prompt]
Image 1: <effect reference>
Image 2: <candidate edit>
<reference-grounded yes/no question> Answer with Yes or No.

Target answer: Yes
\end{lstlisting}

For example, one snow criterion is: ``Does the snow in Image 2 cover the
rooftops and lawns as it does in Image 1?'' Each request contains three such
criteria, which remain fixed during optimization.

\clearpage
\subsection{Frozen Gemini 2.5 Flash Score Judge}

The absolute Gemini 2.5 Flash evaluator~\citep{googledeepmind2025gemini25} receives images in the order source, edited
candidate, effect reference. Its system instruction is:

\begin{lstlisting}[style=prompt]
You are a meticulous, impartial image-editing evaluator. You are shown an EDITED image produced from a SOURCE image by an edit instruction, sometimes the SOURCE image itself, and sometimes a REFERENCE image that exemplifies the desired effect or style. You rate three dimensions on a 0-10 scale (0 = terrible, 10 = perfect). Be calibrated and use the full range. Respond with ONLY a JSON object, no prose, no markdown fences.
\end{lstlisting}

For the three-image reference-conditioned case, the user message is:

\begin{lstlisting}[style=prompt]
Edit instruction: '<instruction>'
There are 3 image(s): the FIRST image is the SOURCE; the SECOND image is the EDITED image; the THIRD image is the REFERENCE.

Judge source_consistency by how faithfully regions NOT targeted by the instruction are preserved from the SOURCE.

A REFERENCE image is provided: it is an EXEMPLAR of the desired effect/style on a DIFFERENT subject (NOT a pixel-exact target). Judge edit_adherence by how well the EDITED image achieves the SAME effect/style shown in the REFERENCE while following the instruction - do NOT penalize differences in subject, scene, or exact pixels between the edited image and the reference.

Rate three dimensions, each an integer or decimal from 0 to 10:
- perceptual_quality: photorealism and absence of artifacts/blur/distortion in the edited image.
- source_consistency: how faithfully regions NOT targeted by the instruction are preserved.
- edit_adherence: how well the intended edit was actually implemented (per the note above).

Return JSON exactly: {"perceptual_quality": <number>, "source_consistency": <number>, "edit_adherence": <number>}
\end{lstlisting}

The Gemini score in Table~\ref{tab:main} is the mean of perceptual quality
and source consistency. The prompt also returns an edit-adherence score,
which is not included in this aggregate.

\clearpage
\section{User Study}
\label{app:user_study}

Each blinded survey item used a $2\times2$ comparison video. The source and effect reference remained static in the top row; the bottom row showed anonymous Result A and Result B. Synchronized slider animations swept between the source and each method's final output. Method assignment to A and B was counterbalanced equally across cases and seeds, and item order was shuffled across effect families; model names were hidden from participants.

The survey also included two rain cases. After data collection, we excluded rain from the analysis because this category contained only two requests in the dataset. The reported analysis includes 16 cases from the 81 requests across eight effect families, two per effect family. Each case was compared separately with Nano Banana 2 and Step1X-Edit v1.2. The reported comparisons use participants' A, B, or ``Hard to tell / No clear winner'' choices for source preservation and effect transfer. The survey called the first criterion ``identity of the input image'' and explicitly described it as camera angle, scene layout, and subjects. We report this criterion as source preservation throughout the paper. The instructions distinguished unintended source changes from changes caused by the requested effect. Most respondents were not VFX experts.

\begin{figure}[H]
    \centering
    \includegraphics[page=1,width=\textwidth]{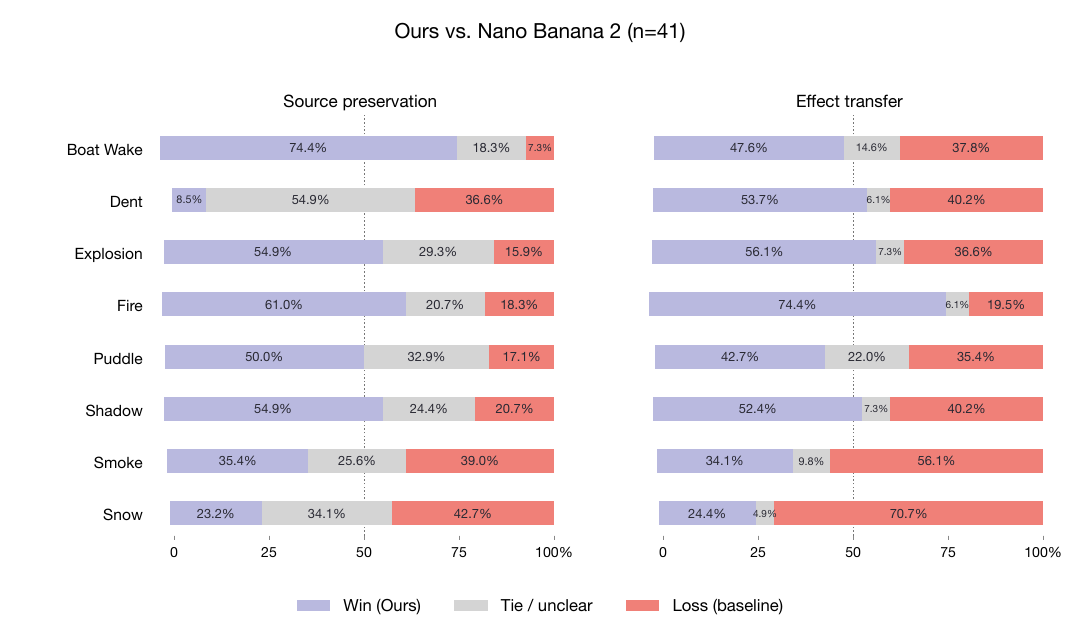}
    \caption{Preference by effect family for source preservation and effect transfer against Nano Banana 2, using 41 responses. Purple favors \textsc{Kinesys}; gray denotes a tie or unclear choice; \mbox{red favors the baseline.}}
    \label{fig:user_study_by_effect}
\end{figure}

\begin{figure}[H]
    \centering
    \includegraphics[page=2,width=\textwidth]{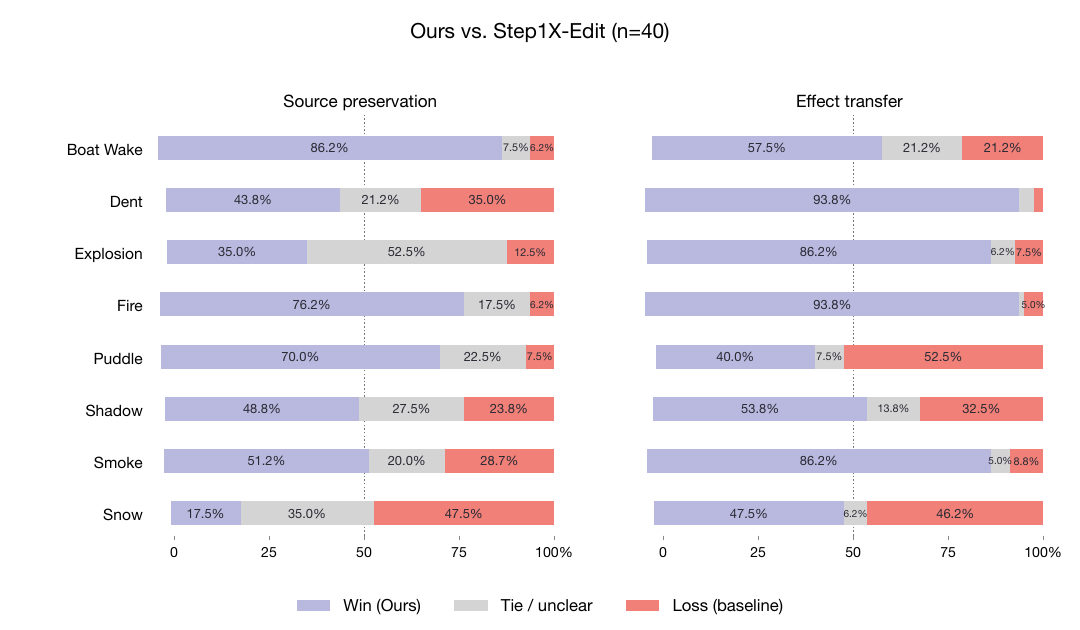}
    \caption{Preference by effect family for source preservation and effect transfer against Step1X-Edit v1.2, using 40 responses. Colors follow Figure~\ref{fig:user_study_by_effect}.}
    \label{fig:user_study_by_effect_step1x}
\end{figure}

Figures~\ref{fig:user_study_by_effect} and~\ref{fig:user_study_by_effect_step1x} keep ties visible and break the source preservation and effect-transfer answers down by family. The Nano Banana 2 comparison contains 41 completed response sets and the Step1X-Edit comparison contains 40. Each set contributes judgments on all 16 comparisons. With two cases per family, each set contributes two choices to each family bar for each question. The main-paper chart averages over these eight families.

\begin{figure}[H]
    \centering
    \includegraphics[width=0.80\textwidth]{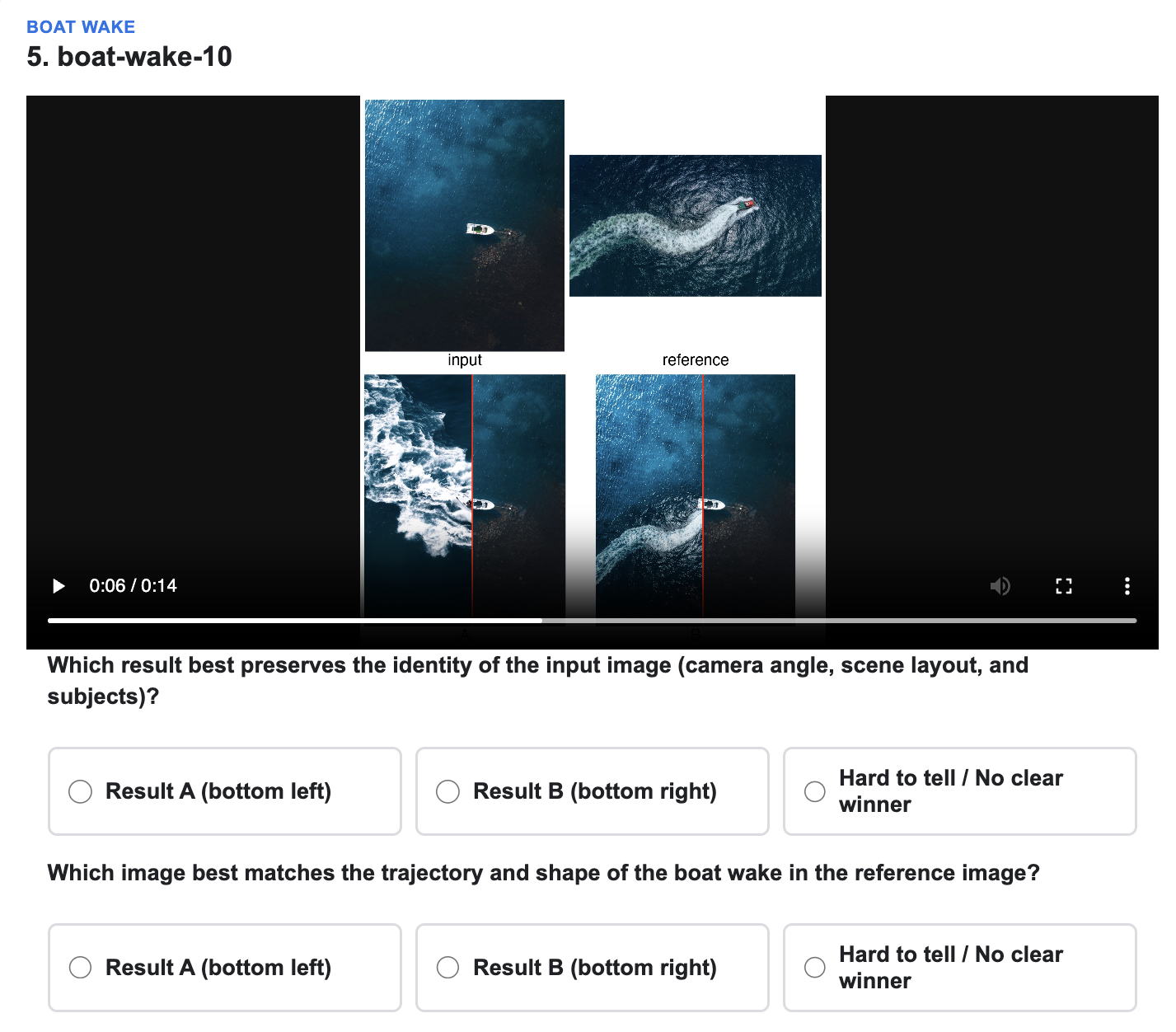}
    \caption{Example survey question for the boat-wake effect, showing the video and original wording. The first question corresponds to source preservation.}
    \label{fig:user_study_question}
\end{figure}

\end{document}